# Dipole Vectors in Images Processing

A. Sparavigna
Dipartimento di Fisica, Politecnico di Torino, Torino, Italy
amelia.sparavigna@polito.it

**Abstract**: Instead of evaluating the gradient field of the brightness map of an image, we propose the use of dipole vectors. This approach is obtained by adapting to the image gray-tone distribution the definition of the dipole moment of charge distributions. We will show how to evaluate the dipoles and obtain a vector field, which can be a good alternative to the gradient field in pattern recognition.

## 1. Introduction

The gradient field of a continuous function defined over the image frame is used in image processing to investigate the likelihood of images, in particular to see if they are invariant for rotation and translation [1]. Gradient field is also used in the reconstruction of images, in several important applications, from pattern recognition to image reconstruction [2-5].

The gradient is a well-known concept in physics. Let us remember that an electric field can be defined according to the gradient of an electric potential. In fact, we have a scalar quantity, the potential function $V$, from which we obtain a vector field $\vec{E}$, according to the following equation:

$$\vec{E} = -\text{grad}\, V = -\left(\frac{\partial V}{\partial x}\vec{u}_x + \frac{\partial V}{\partial y}\vec{u}_y\right) \tag{1}$$

in the two-dimensional case. $\vec{u}_x, \vec{u}_y$ are the unit vectors of the orthogonal frame.

Using the gray-tone map of an image as a potential function, the gradient of it can be evaluated and the corresponding vector field obtained.

In this paper we propose another vector field that can be easily evaluated from the same image map. In Ref.[6], we started a discussion on the role of dipoles and quadrupoles for image processing. We showed that dipole and quadrupole moments, suitably defined for image maps, can be used for the image edge detection. In particular we used the magnitude of dipoles in the algorithm for edge detection. Here, we want to discuss dipole moments as quantities able to give a vector field which can be a good substitute of the gradient field.

A possible application in the field of pattern recognition is also proposed.

## 2. Dipole moments for images.

A general distribution of electric charges can be described by its net charge, by its dipole and quadrupole moments, and higher order moments [7]. The dipole moment is a vector, the direction of which is toward the positive charge.

Consider a collection of $N$ particles with charges $q_n$ and position vectors $\vec{r}_n$. The physical quantity defining the dipole vector is given by:

$$\vec{p} = \sum_{n=1}^{N} q_n \vec{r}_n \qquad (2)$$

For a distribution of charges in a plane $(x,y)$:

$$p_x = \sum_{n=1}^{N} q_n x_n \;\; ; \;\; p_y = \sum_{n=1}^{N} q_n y_n \qquad (3)$$

where $x_n, y_n$ are the Cartesian components of the position vector of each charge. We are assuming the distribution of charges in a $(x,y)$-plane, because we want to apply these moments to a two-dimensional image. In the case of images, we have a quantity which can play the role of a charge distribution and this quantity is the image bitmap $b$, which we consider for simplicity as a gray-tone map. The bitmap representation of an image consists of a function which yields the brightness of each point within a specific width and height range:

$$b: \; D \to B \qquad (4)$$

with $D = I_h \times I_w$, where $I_h = \{1,2,\ldots,h\} \subset \mathrm{N}$, $I_w = \{1,2,\ldots,w\} \subset \mathrm{N}$ and $B = \{0,1,\ldots,255\} \subset \mathrm{N}$.

We can evaluate a dipole moment for the overall image in the following way [6]:

$$p_x = \frac{1}{hw} \sum_{i=1}^{h} \sum_{j=1}^{w} b(i,j)\, x_{ij} \; ; \; p_y = \frac{1}{hw} \sum_{i=1}^{h} \sum_{j=1}^{w} b(i,j)\, y_{ij} \qquad (5)$$

where $x_{ij}, y_{ij}$ are the coordinates of pixel at position $(i,j)$ in the image frame. These coordinates can be simply identified with the pixel positions: $x_{ij} = i, y_{ij} = j$.

To consider the whole image is of course not interesting. Let us evaluate the dipole moment on a neighborhood of each pixel. As usual, the pixel position in the image map is given by indices $(i,j)$. The neighborhood consists of all pixels with indices contained in the two following intervals $I_i = [i - \Delta i, i + \Delta i]$ and $I_j = [j - \Delta j, j + \Delta j]$. The local average brightness is defined as:

$$M(i,j) = \frac{1}{4\Delta i \, \Delta j} \sum_{k \in I_i} \sum_{l \in I_j} b(k,l) \qquad (6)$$

A pixel in this local neighborhood can have a "charge", that can be positive or negative, if we define the "charge" as $q(i,j) = b(i,j) - M(i,j)$. Local dipoles are then given by:

$$p_x(i,j) = \frac{1}{4\Delta i \, \Delta j} \sum_{k \in I_i} \sum_{l \in I_j} q(k,l) x_{kl}$$

$$p_y(i,j) = \frac{1}{4\Delta i \, \Delta j} \sum_{k \in I_i} \sum_{l \in I_j} q(k,l) y_{kl} \qquad (7)$$

These are the components of the image dipole moment. With Eqs.7 we can obtain the vector field $\vec{p}(i,j) = p_x(i,j)\vec{u}_x + p_y(i,j)\vec{u}_y$. We can prepare a image map considering just the magnitude of dipole vectors. This is simply a scalar field given by:

$$P(i,j) = \left( p_x^2(i,j) + p_y^2(i,j) \right)^{1/2} \qquad (8)$$

Let us evaluate the maximum value of $P(i,j)$ in the total image frame and call it $P_{Max}$. To represent the field of the magnitudes of dipoles, we associate with each pixel a gray tone as follows:

$$b_P(i,j) = 255 \left( \frac{P(i,j)}{P_{Max}} \right)^\alpha \qquad (9)$$

where exponents $\alpha$ is adjusted to enhance the visibility of the map. A good choice is $\alpha = 1/2$.

The image dipole moments describe the distribution of tones. We tested in [6], the image dipole moments for the edge detection. The results are comparable with those obtained applying GIMP software. Let us remember that an edge corresponds to a change of brightness in a small region of the image frame. Applying an edge detector to an image map gives a set of curves describing the boundaries of objects.

Let us apply the dipole approach to an image with hieroglyphic signs. The dipole moment is evaluated on the smallest possible neighborhood, that is the neighborhood with $2 \times 2$ pixels. Fig.1 shows the original image and the map representing the magnitude of local dipole moments. This map enhances the contours of the hieroglyphic writing.

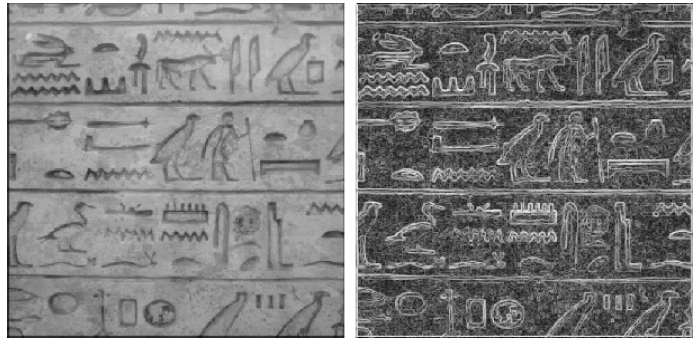

Fig.1 The original image of hieroglyphic writing on the left. On the right, the map obtained evaluating the magnitude of local dipole moments, according to Eqs.8 and 9. Note that the map is clearly enhancing the contours of the writing.

**3. The vector field of dipole directions.**
To illustrate the vector characteristic of dipole moments, let us apply the image dipole moments calculation to an image of Brodatz album [8]. The evaluation of the magnitude gives the edge detection as in Fig.2.

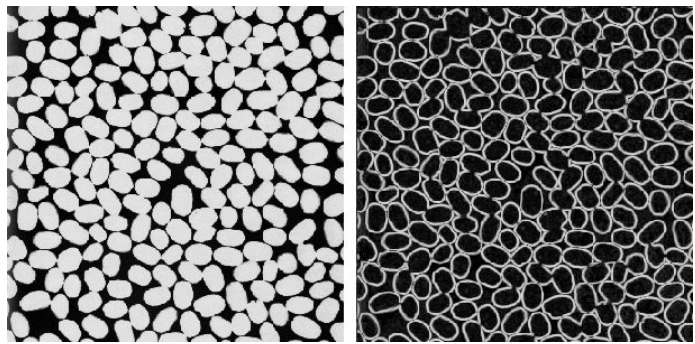

Fig.3 The original image map on the left from the Brodatz album. On the right, the map obtained evaluating the dipole moments, according to Eqs.8 and 9.

What is the behavior of the directions of dipole moments? This is not easy to represent with a map. For simplicity, let us consider the image subdivided in squares of $20 \times 20$ pixels. In each square, a red line is giving the direction of the dipole. The dipole moment is evaluated on the same neighborhood.

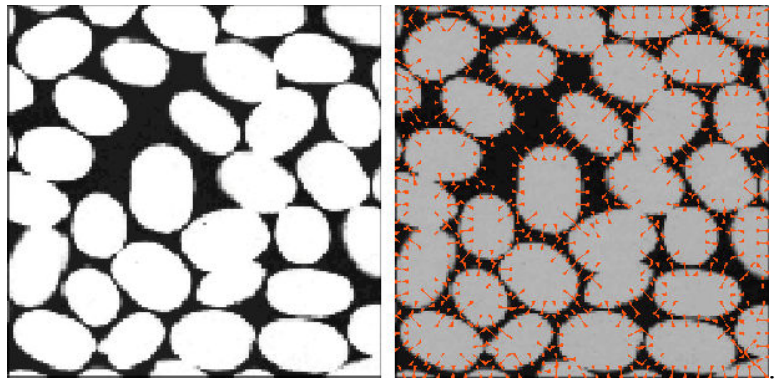

Fig.2 On the right, the behavior of dipole directions. The dipole vectors are represented by red lines. For simplicity, the image is subdivided in squares of $20\times 20$ pixels. The dipoles with a magnitude lower than a certain threshold value are not shown.

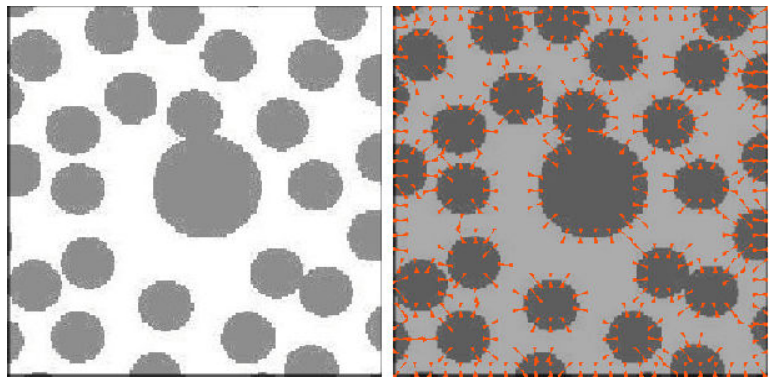

Fig.3 Another example of the behavior of dipole directions. The image is subdivided in squares of $20\times 20$ pixels. Note that the red lines, that is the dipoles, are perpendicular to the boundaries.

In Figs. 2 and 3 we used a very rough representation of the vector field. More sophisticated drawings are possible with LIC or TOSL (see for instance Refs. 9 and 10). The Figures are proposed just to show that the red lines are almost perpendicular to the edges of the image. Of course, the vector field is more precise when the neighborhood is reduced.
Note that we are not evaluating the gradient of the tone map, we are evaluating the dipole moments. Moreover, the dipole moment is crossing the boundary between regions with positive and negative charges. As the magnitude of the dipole is able to give the edge, that is the boundary between the regions, the direction of the dipole is giving the direction perpendicular to the boundary.
We are studying the possibility of using the vector field of dipole directions, in those applications where the gradient field is involved. Another interesting application is in the field of signs recognition. In fact, choosing a proper threshold for the magnitude of dipoles, we can separate the signs from the background. For each sign we can evaluate the vector field of dipole moments (see for instance Fig.4 with hieroglyphic sings). Once we have this vector field, we can evaluate another interesting field. This field has vectors which have directions perpendicular to the dipoles.

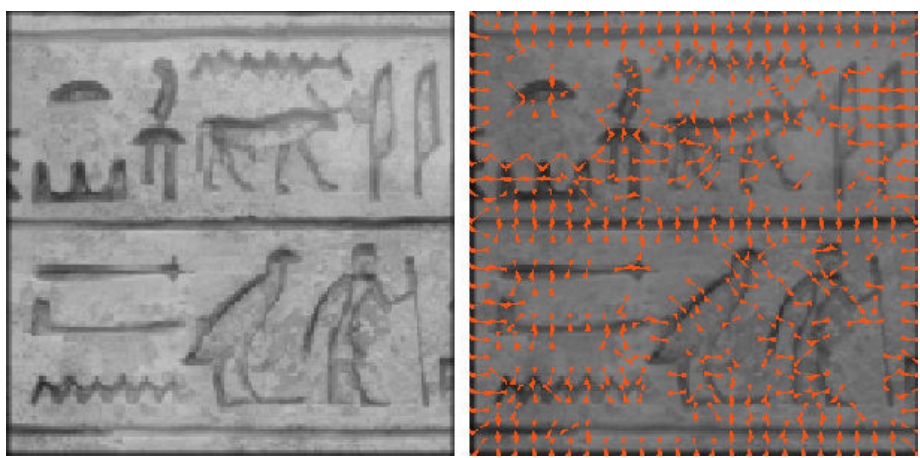

Fig.4 Choosing a proper threshold for the magnitude of dipoles, the dipole moments can be used in separating the image in many domains, each of them containing a sign. The red lines show the dipoles

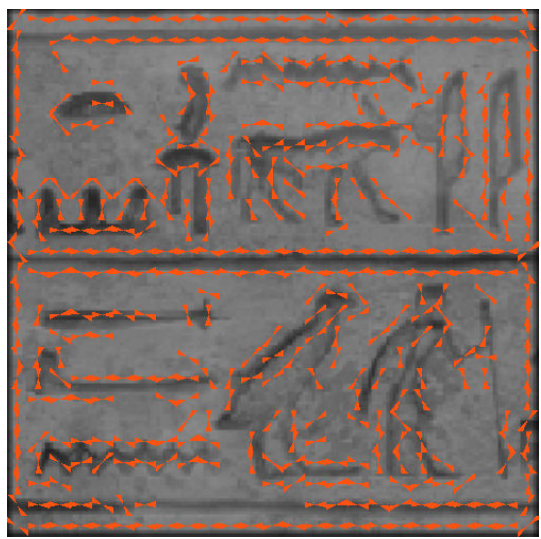

Fig.5 It is also possible to evaluate the field given by the directions perpendicular to the dipole vectors. This field can be used to recognize the signs.

Fig.5 shows the vectors perpendicular to dipole vectors in Fig.4. These vectors describe a field, the lines of which are perpendicular to lines of the dipole field. This field can be used for pattern recognition. In fact, Fig.5 can be considered as a preliminary example of the use of dipoles in recognition of hieroglyphic sings.

**5. Conclusions**

The paper describes an algorithm based on image dipole moment. The moment is defined as in physics is defined the dipole moment of a charge distribution. With this algorithm, evaluating the moments on small neighborhoods of each pixel, it is possible to detect the edges in the image frame by means of the magnitude of the dipole. Moreover it is possible to define a vector field, able to substitute the gradient field in pattern recognition. Algorithms for this purpose are under development.